\documentclass[conference]{IEEEtran}

\usepackage{diagbox}
\usepackage{graphicx}
\usepackage{textcomp}
\usepackage{xcolor}
\usepackage[linesnumbered,ruled,vlined]{algorithm2e}
\usepackage{subfigure}
\usepackage{booktabs}
\usepackage{amsmath}
\usepackage{multirow}
\usepackage{makecell}

\IEEEoverridecommandlockouts
\usepackage{cite}
\usepackage{amsmath,amssymb,amsfonts}
\usepackage{algorithmic}
\usepackage{graphicx}
\usepackage{textcomp}
\usepackage{xcolor}
\def\BibTeX{{\rm B\kern-.05em{\sc i\kern-.025em b}\kern-.08em
    T\kern-.1667em\lower.7ex\hbox{E}\kern-.125emX}}
\begin{document}

\title{Generating Image Adversarial Examples by Embedding Digital Watermarks\\
\thanks{The first two authors contributed equally to this work.}
}

\author{\IEEEauthorblockN{Yuexin Xiang}
\IEEEauthorblockA{\textit{School of Computer Science} \\
\textit{China University of Geosciences}\\
Wuhan, China \\
yuexin.xiang@cug.edu.cn}
\and
\IEEEauthorblockN{Tiantian Li}
\IEEEauthorblockA{\textit{Faculty of Engineering and Information Technology} \\
\textit{University of Melbourne}\\
Melbourne, Australia \\
tiantian3@student.unimelb.edu.au}
\and
\IEEEauthorblockN{Wei Ren}
\IEEEauthorblockA{\textit{School of Computer Science} \\
\textit{China University of Geosciences}\\
Wuhan, China \\
weirencs@cug.edu.cn}
\and
\IEEEauthorblockN{Tianqing Zhu}
\IEEEauthorblockA{\textit{School of Computer Science} \\
\textit{China University of Geosciences}\\
Wuhan, China \\
tianqing.zhu@ieee.org}
\and
\IEEEauthorblockN{Kim-Kwang Raymond Choo}
\IEEEauthorblockA{\textit{Department of Information Systems and Cyber Security} \\
\textit{University of Texas at San Antonio}\\
San Antonio, USA \\
email address or ORCID}
}

\maketitle

\begin{abstract}
With the increasing attention to deep neural network (DNN) models, attacks are also upcoming for such models. For example, an attacker may carefully construct images in specific ways (also referred to as adversarial examples) aiming to mislead the DNN models to output incorrect classification results. Similarly, many efforts are proposed to detect and mitigate adversarial examples, usually for certain dedicated attacks. In this paper, we propose a novel digital watermark-based method to generate image adversarial examples to fool DNN models. Specifically, partial main features of the watermark image are embedded into the host image almost invisibly, aiming to tamper with and damage the recognition capabilities of the DNN models. We devise an efficient mechanism to select host images and watermark images and utilize the improved discrete wavelet transform (DWT) based Patchwork watermarking algorithm with a set of valid hyperparameters to embed digital watermarks from the watermark image dataset into original images for generating image adversarial examples. The experimental results illustrate that the attack success rate on common DNN models can reach an average of 95.47\% on the CIFAR-10 dataset and the highest at 98.71\%. Besides, our scheme is able to generate a large number of adversarial examples efficiently, concretely, an average of 1.17 seconds for completing the attacks on each image on the CIFAR-10 dataset. In addition, we design a baseline experiment using the watermark images generated by Gaussian noise as the watermark image dataset that also displays the effectiveness of our scheme. Similarly, we also propose the modified discrete cosine transform (DCT) based Patchwork watermarking algorithm. To ensure repeatability and reproducibility, the source code is available on GitHub.
\end{abstract}


\section{Introduction}
\label{sec:introduction}

Deep neural networks (DNNs) are increasingly been used in image recognition for many different applications such as monitoring of crowds, surveillance at key installations, and clinical diagnosis. Similar to any technology, people will attempt to find vulnerabilities and weaknesses in the technologies in order to carry out malicious or nefarious activities. Szegedy et al. \cite{szegedy2013intriguing} revealed that minor perturbation in the image could lead to incorrect identification results in deep neural networks, and introduced the concept of adversarial examples. The latter referred to images that are purposefully crafted to result in misclassification of deep neural networks, for example, due to perturbations.  

Since the seminal work of Szegedy et al. \cite{szegedy2013intriguing}, many other researchers have proposed different schemes to generate adversarial examples, such as those using gradient variations in deep neural networks or elaborate perturbations on images to generate adversarial examples. Such approaches can also be categorized into white-box approaches and black-box approaches. Examples of white-box attack approaches include the fast gradient sign method (FGSM) \cite{goodfellow2014explaining} and the DeepFool method \cite{moosavi2016deepfool}, both of which are based on gradient design. Some approaches (e.g., Jacobian-based saliency map attack (JSMA) method \cite{wiyatno2018maximal}) attempt to reduce the number of modified pixels in an image with minimal interference to the entire image. Examples of black-box approaches include one-pixel attack and local search attack \cite{narodytska2016simple}, both of which do not require the attacker to know about the parameters and gradient changes in the targeted deep neural networks.

Existing popular methods for generating adversarial examples generally focus on using gradient variations and crafty perturbations and ensure subtle perturbations by limiting their $L_p$ norm. We take a different approach that focuses on the  method itself in this paper, as described below:

\begin{enumerate}
\item We propose an improved Patchwork watermarking algorithm based on discrete wavelet transform (DWT) with a set of hyperparameters for efficiently generating image adversarial examples, the source code is available on Github\footnote{https://github.com/Y-Xiang-hub/Generating-Image-Adversarial-Examples-by-Embedding-Digital-Watermarks
}.

\item We propose a novel watermarking algorithm-based framework to generate adversarial examples, which consists of three essential modules and two kinds of datasets.

\item We can see from the experimental results that the success rate can achieve an average of 95.47\% on the CIFAR-10 dataset while the processing speed of each image is an average of 1.17 seconds.

\item We also propose the modified discrete cosine transform (DCT) based Patchwork watermarking algorithm. 

% \item We provide a set of embedding parameters with excellent performance through large amounts of experiments to generate adversarial examples. We also propose a scheme that uses the confidence of the host images to select watermark images, with the aim of making the experiment more efficient.
% \item We explore whether the effectiveness of our scheme can be enhanced if we use the features extracted from watermark as the watermark. The findings suggest that using features as watermarks can result in a higher success rate and reduce the perturbations of embedding.
\end{enumerate}

\section{Related Work}
\subsection{Adversarial Example Generation}
The security of artificial intelligence systems, especially DNN-based systems, has attracted the attention of the research community and industrial areas, which is mainly caused by a number of methods that appeared to generate adversarial examples for attacking deep neural networks in recent years. For instance, Goodfellow et al. \cite{goodfellow2014explaining} suggested FGSM can be used as directed attacks and undirected attacks against deep neural networks based on gradient. DeepFool \cite{moosavi2016deepfool} was also utilized as an undirected attack, which results in fewer perturbations compared with FGSM according to their analysis. Narodytska et al. \cite{narodytska2016simple} proposed the single pixel attack, which can mislead deep neural networks by only changing one pixel on the image. Moreover, the authors improved their method as the local search attack, which works better than the single pixel attack when the total number of pixels of an image is larger.

There are several researchers who design or apply novel frameworks to generate adversarial examples. Baluja et al. \cite{baluja2017adversarial} proposed the adversarial transformation networks (ATNs) to promptly generate adversarial examples using the self-supervisor forward propagation neural networks. Such a method can provide a variety of different types of adversarial examples, targeting deep neural networks. Chen et al. \cite{chen2017zoo} developed a black-box attack based on zeroth order optimization (ZOO) to generate adversarial examples by straightly reckoning the gradients of the selected DNN only according to its input and output, they also proved the effectiveness of their scheme by experiments. Xiao et al. \cite{xiao2018generating} presented a model based on the generative adversarial networks (GANs) to generate high-quality adversarial examples in a relatively short time. The generator that successfully passes the test of the discriminator can produce the special adversarial examples that the target model training by the original data cannot rightly classify. 

Besides, other researchers attempt to attack DNNs from other particular perspectives. Deng et al. \cite{deng2019generate} proposed a spatial transformed attack method based on attention, which can be used to search for significative areas to be targeted. Such areas can also be subjected to spatial transformation. The approach is generally attack-efficient while decreasing the perturbations. In another separate work, Qiu et al. \cite{qiu2019semanticadv} proposed SemanticAdv for generating a new genre of semantically realistic adversarial examples through attribute-conditioned image editing. Their approach has a high success rate on both black-box attacks and white-box attacks, which can circumvent detection approaches based on pixels and attributes. Tu et al. \cite{tu2019autozoom} proposed a universal framework that gives solutions to attacking DNNs under black-box circumstances called Autozoom. This scheme not only successfully improves the effectiveness of finding the adversarial examples, but also greatly decreasing the average query counts during the process.

Moreover, Chen et al. \cite{chen2020hopskipjumpattack} proposed HopSkipJumpAttack that can generate adversarial examples only on the basis of the output labels provided by the targeted DNN. This method applies the binary information at the decision boundary for estimating the gradient direction. Qiu et al. \cite{qiu2020semanticadv} proposed SemanticAdv that can conduct adversarial examples by attribute-conditioned image editing. The experiments under black-box and white-box settings show high attack success rates. Also, this attack is hard to be defended through normal defense methods. Croce and Hein \cite{croce2020minimally} proposed a novel way to generate adversarial examples under the white-box condition that has an intuitive geometric meaning. In addition, this method is able to minimize the range of perturbation and produce high-quality adversarial examples. 

Additionally, Wu et al. \cite{wu2020skip} found some DNNs (e.g. ResNet and DenseNet) have security weaknesses caused by one of the important components in DNN - skip connections. Through the weakness of this structure, Wu et al. \cite{wu2020skip} proposed Skip Gradient Method (SGM) to conduct adversarial examples more efficiently. However, most adversarial examples cannot show their strength in the real physical world, thus, Kong et al. \cite{kong2020physgan} proposed a method for physical-world-resilient adversarial example generations called Physgan. Adversarial examples generated by Physgan to attack common autonomous driving scenarios, and the experiments illustrate the effectiveness and the robustness of the scheme. Besides, Yang et al. \cite{yang2021attacking} presented a group gradient-based approach to generate adversarial examples in the real scenario aiming at attacking deep learning-based layout hotspot detectors. Recently, Xiang et al. \cite{xiang2021local} proposed a black-box attack that only produces perturbations on the key area of the image-based within limited queries. This framework is built through two vital parts respectively called the transferability of model interpretations and transferability of adversarial examples.

\subsection{Digital Watermark Application in DNN}
% \subsubsection{Digital Watermark in DNN}

Digital watermarking has also been applied to DNNs, such as protecting DNNs (e.g., see the scheme of Zhang et al. \cite{zhang2018protecting} for DNNs' intellectual property protection). In another work, Le et al.\cite{le2019adversarial} proposed a remote watermarking extraction scheme to ensure flexibility in watermarking-enabled DNN protection. DNNs can also be applied in the watermarking domain. For example, Quiring et al. \cite{quiring2018adversarial} used DNNs to extract digital watermarks in images without knowing the underpinning watermarking scheme, additionally, Wen et al. \cite{wen2019romark} proposed using DNNs to build a robust watermarking system.

% \subsubsection{Digital Watermark in Adversarial Example for DNN}

Another research direction is to utilize digital watermarks to mislead the DNNs. For example, Shafahi et al. \cite{shafahi2018poison} proposed using watermarking to facilitate poisoning attacks against DNNs. Although most of the adversarial example attacks using digital watermarks focused on images, it has been suggested that watermarking can also be utilized in other types of DNNs. For example, Chen et al. \cite{chen2020attacking} proposed a scheme to utilize digital watermarks in text, so that text recognition produces incorrect results. Jia et al. \cite{jia2020adv} proposed Adv-watermark that putting the visible watermarks on the images to cause perturbations for generating adversarial examples. They also proposed a new optimization algorithm named Basin Hopping Evolution (BHE) that is combined with Adv-watermark to generate adversarial examples under black-box settings in a short time.

For invisible digital watermark-based methods, Anshumaan et al. \cite{anshumaan2020wavetransform} proposed WaveTransform that can generate adversarial noises according to the low-frequency and high-frequency subbands. Specifically,  WaveTransform applies wavelet decomposition to analyze the frequency subbands, then the frequency subbands are also used for conducting the adversarial examples. Yahya et al. \cite{yahya2020probabilistic} presented applying frequency-domain approaches, such as DWT and DCT, to craft perturbations in frequency component for generating adversarial examples by embedding a special designed digital watermark image to the targeted images from the MINST dataset. 

% \subsection{Current Issue}

However, for the scheme proposed by Anshumaan et al. \cite{anshumaan2020wavetransform}, they only adopt wavelet decomposition to explain how to generate suitable noise and did not consider applying embedding digital watermarks to generate adversarial examples. In addition, for the work of Yahya et al. \cite{yahya2020probabilistic}, though they attempt to embed a special watermark image to generate adversarial examples for the MINST dataset, the generation process of this special watermark is not clear and the scheme is hard to extend to other datasets or more DNN structures.

Building on the existing literature, we explore the potential of generating adversarial examples based on the workings of DNNs. The basic principle of DNNs for recognizing images is to extract the features of the images layer by layer, embedding digital watermark information into the image means that features that are difficult to recognize by human vision are somehow added to the targeted image. That is possible to interfere with the image recognition process of DNNs. We will introduce the work related to the watermarking algorithm in the next section.

\section{Preliminary}
Bender et al. \cite{bender1996techniques} proposed a statistical method called Patchwork to hide information. They adjust the brightness of the image to achieve information hiding, at the same time, they keep the balance of the overall brightness of the image that humans are hard to recognize the changes. However, using Patchwork to embed watermarks by changing the image's statistical characteristics only hides a very small amount of information in the image.

In order to increase the capacity of hidden information, Mei et al. \cite{jiansheng2009digital} proposed a method based on DWT and DCT for embedding the image type digital watermark into the host image. In other words, their method transforms the host image with the DWT while using the two-dimensional DCT to transform the watermark image. Finally, the result of the watermark image transformation is added to the high-frequency component of the host image transformation.

The weaknesses of the schemes proposed by Anshumaan et al. \cite{anshumaan2020wavetransform} and Yahya et al. \cite{yahya2020probabilistic} and the literature mentioned in this section motivates us to explore the possibility of the combination of the thought of Patchwork and DWT-based and DCT-based watermarking algorithms. In addition, the improved combination algorithms are able to be applied to generate adversarial examples in DNNs.

According to the work of Zhou et al. \cite{zhou2009blind} and Zhang et al. \cite{5364628} the equation for the sensitivity of human visual system (HVS) to RGB primary colors: 
\begin{equation}\label{HVS}
y=0.299R+0.587G+0.144B,
\end{equation}
where \emph{y} is the luminance; \emph{R} is red component; \emph{G} is green component; \emph{B} is blue component. We use Eq.(\ref{HVS}) to keep the luminance balance of the targeted image while image type data (i.e. RGB images) can be embedded into the targeted image as the watermark by DWT-based watermarking algorithm for generating adversarial examples. The specific algorithm and framework will be shown in the next section.

\section{Proposed Scheme}

Notations that are high-frequency use in this paper are listed in Table \ref{tab-nota}.
\begin{table}[!htbp]
\setlength{\belowcaptionskip}{+0.1cm}
\caption{Notations} \label{tab-nota} \centering
\begin{tabular}{l p{6.5cm}}
\noalign{\hrule height 0.5pt}
Notations & Description\\
\noalign{\hrule height 0.5pt}
$C_i$ & the confidence of the image\\ 
$C_{i(N)}$ & the confidence of the image after embedding the watermark image for \emph{N} number of times\\
$T_i$ & the type of the image\\
$T_{i(N)}$ & the type of the image after embedding the watermark image for \emph{N} number of times\\
$I_i$ & the image\\ 
$I_{wmi}$ & the watermark image\\
$I_{i(N)}$ & the image that the watermark image has embedded for \emph{N} number of times\\
$N$ & the number of times of embedding the watermark\\
$N_{max}$ & the maximum number of times of embedding the watermark\\
$S_r$ & the watermark embedding strength of red component\\
$S_g$ & the watermark embedding strength of green component\\
$S_b$ & the watermark embedding strength of blue component\\
\noalign{\hrule height 0.5pt}
\end{tabular}
\end{table}

\subsection{Framework}
\begin{figure*}[!htbp]
  \centering
  \includegraphics[width=0.8\linewidth]{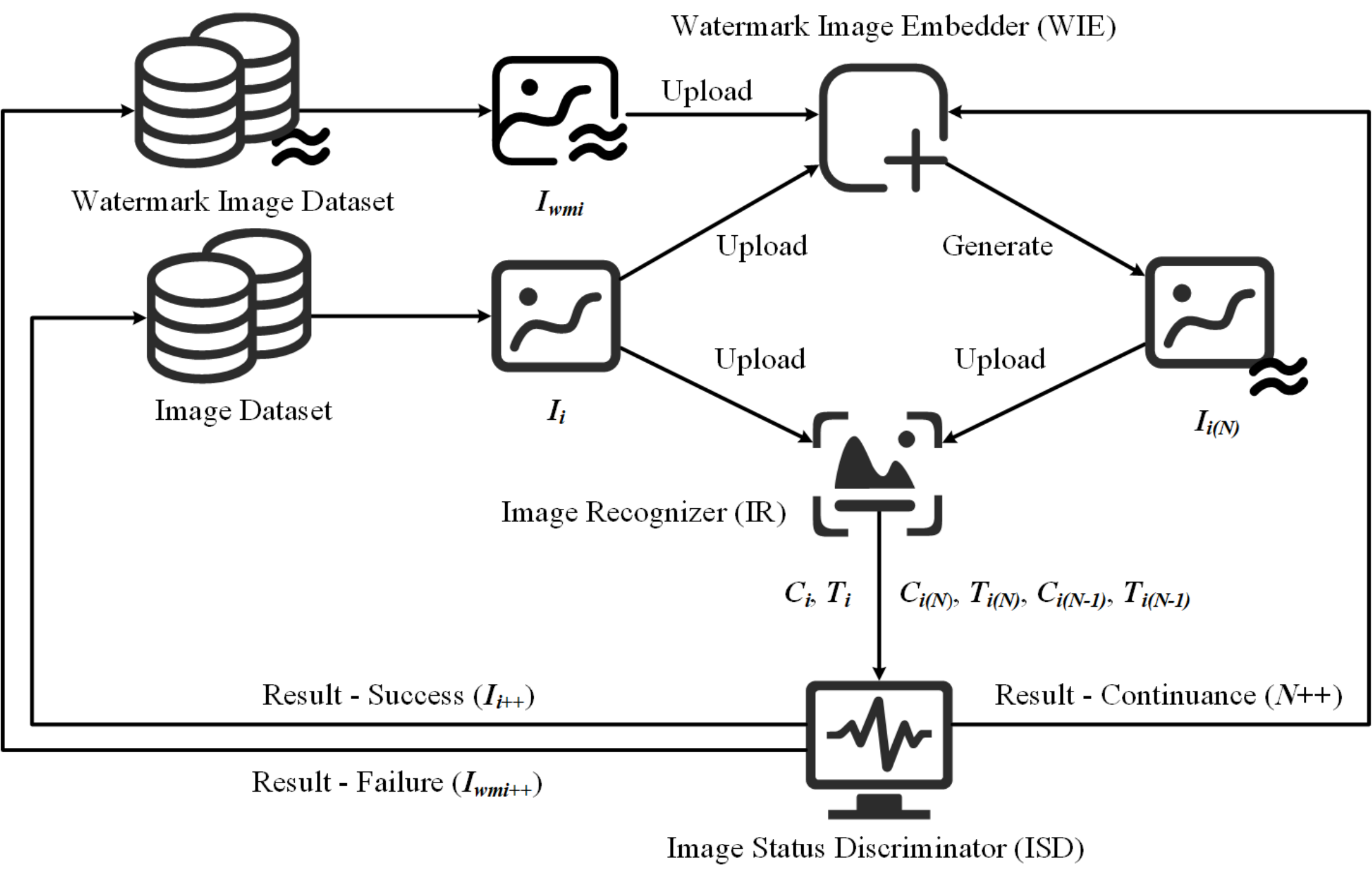}
  \caption{Framework of proposed scheme}
  \label{framework}
\end{figure*}

The framework of the proposed scheme is shown in Fig.\ref{framework} that consists of three vital modules, respectively the image recognizer (IR), the watermark image embedder (WIE), and the image status discriminator (ISD). 

In a nutshell, We process each image in the image dataset one by one, concretely, for each image from the image dataset, we randomly select 1\% watermark images from the whole watermark dataset. Then we embed each selected watermark image using the WIE module with a finite number of times and specific embedding strength into each image from the image dataset to generate a new image. 

We start to process the next image from the image dataset until the type of newly generated image is found incorrect by the ISD module according to the results uploaded from the IR module (i.e., this new image is an adversarial example) or the process exceeds the number of times limit. We will introduce these three parts in detail below.

\subsection{Image Recognizer}
In the IR module, the IR module recognizes the uploaded $I_i$ and records the results $C_i$ and $T_i$ of $I_i$, then IR uploads them to the ISD module. In addition, after the WIE module embeds $I_{wmi}$ $N(1 \leq N \leq N_{max})$ times into $I_i$, the generated $I_{i(N)}$ will upload to the IR module and IR recognizes $I_{i(N)}$ and records the results $C_{i(N)}$ and $T_{i(N)}$ of $I_{i(N)}$, then IR also uploads them to the ISD module. 

After each round embedding for an image from the image dataset terminates, the IR module uploads the results of this round to the ISD module for further comparison and analysis. Also, the feedback from the ISD module decides the next operations for the IR module and the WIE module. Besides, according to the requirement of testing different DNN models, various trained DNN models can be applied in the IR module as the function of computing the confidence, such as VGG\cite{VGG} and ResNet\cite{ResNet}.

\subsection{Watermark Image Embedder}
In the WIE module, $I_{wmi}$ will be embedded into $I_i$ for $N$ times by the watermarking algorithm, where $N$ is decided on the parameter settings of the ISD module. Additionally, referring to Eq.\ref{HVS}, we set the default embedding strength parameter proportion of RGB as:
\begin{equation}\label{RGB_P}
    S_r:S_g:S_b = 2:4:1.
\end{equation}

Based on Eq.\ref{RGB_P} and a huge amount of our experimental results, we set the initial embedding strength parameters of RGB for the improved DWT-based watermarking algorithm and the maximum number of times of embedding the watermark as:

\begin{equation}\label{RGB_P_DWT}
    \begin{cases}
    S_r:S_g:S_b = 0.02:0.04:0.01, \\
    N_{max} = 2.
    \end{cases}
\end{equation}

The specific algorithm for generating adversarial examples is described in Algorithm \ref{DWT} in detail. Moreover, considering the flexibility of the proposed framework, we can substitute the current watermarking algorithm applied in the WIE module with any other watermarking algorithm.

% The specific algorithms  are shown in Algorithm \ref{alg:alg1} and Algorithm \ref{alg:alg2}.

%  to set the appropriate  
% and then embed the watermark image into the host image to generate the adversarial example. Before starting to embed the watermarked image into the host image through the watermarking algorithm, two parameters need to be set, namely: $Embed_{s}$ and $Embed_{t}$.

% To generate adversarial examples using the improved discrete wavelet transform based Patchwork algorithm, after setting $Embed_{s}$, we embed the $W_1$ into $H_x$ using this watermarking algorithm. Specifically, we perform a total of ten rounds of embedding $W_1$ into $H_x$, with the first round embedding $W_1$ 5 times into $H_x$, the tenth round embeds 50 times $W_1$ in $H_x$, and five intervals of embedding between rounds. We repeat this process for other selected host images and corresponding $W_i$. In addition, the difference for the modified discrete cosine transform based Patchwork algorithm is that embedding $W_1$ 1 times into $H_x$ in the first round, and embedding 10 times $W_1$ in $H_x$  in the tenth round, and 1 interval of embedding between rounds.

\begin{algorithm}[htbp!]
\DontPrintSemicolon
\KwIn{$I_i$, $I_{wmi}$}
\KwOut{$I_{i(N)}$}
\For {$i = 1, 2, ...$} {
    \For{$j = 1, 2, ..., N$} {
        read and resize $I_{i(j)}$ and $I_{wmi}$;\\

        separate $R$, $G$ and $B$ channels of $I_{i(j)}$ and $I_{wmi}$;\\
        
        \For{$c = \{R, G, B\}$} {
            $cA3^c, cH3^c, cV3^c, cD3^c \leftarrow nDWT(I_{i(j)}^{c}, 3)$;\\
            $cAwm^c \leftarrow nDWT(I_{wmi}^{c}, 1)$;\\
            $EcA3^c \leftarrow cA3^c + S_c \times cAwm^c$;\\
            $EcH3^c \leftarrow cH3^c - S_c \times cAwm^c$;\\
            $EcV3^c \leftarrow cV3^c - S_c \times cAwm^c$;\\
            $EcD3^c \leftarrow cD3^c - S_c \times cAwm^c$;\\
            
            $E^{c} \leftarrow IDWT(EcA3^c, EcH3^c, EcV3^c, EcD3^c)$;\\
        }
        
        $I_{i(j)} \leftarrow cat(3,E^{R},E^{G},E^{B})$;\\
    }
}
\Return $I_{i(j)}$
\caption{Improved DWT-based Watermarking}
\label{DWT}
\end{algorithm}

\subsection{Image Status Discriminator}
In the ISD module, the ISD module receives the parameters uploaded by IR including $C_{i(N)}$, $T_{i(N)}$, $C_{i(N-1)}$, $T_{i(N-1)}$, $C_i$, and $T_i$. Then we can use these parameters to measure whether generated $I_{i(N)}$ is an adversarial example for the specific DNN model as shown in Algorithm \ref{ISD}. We will apply Algorithm \ref{ISD} for each image in the image dataset and finally, compute the success rate.

%----------------------------------------------------------

\begin{algorithm}[htbp!]
\DontPrintSemicolon
\KwIn{$C_{i(N)}$, $T_{i(N)}$, $T_i$}
\KwOut{$Result$, $I_{i(N)}$}
\For{$N \in [1, N_{max}]$}{ 
    \If{$T_{i(N)} = T_i$}{
        $Result \leftarrow Continuance$\\
        $N \leftarrow N + 1$\\
        \emph{call} WIE module and IR module\\
    }
    \Else {
        \If{$C_{i(N)} \textgreater threshold$}{
            $Result \leftarrow Success$;\\
            \Return $Result$, $I_{(i(N))}$\\
        }
        \Else{
            $Result \leftarrow Continuance$\\
            $N \leftarrow N + 1$\\
            \emph{call} WIE module and IR module\\
        }
    }
}
$Result \leftarrow Failure$;\\
\Return $Result$
\caption{ISD Module}
\label{ISD}
\end{algorithm}

%---------------------------------------------------------
Furthermore, in order to decrease the time cost for generating the adversarial examples, when $T_i(N) = T_i(N-1) = T_i$, we can compute $C_{i(N)}-C_{i(N-1)}$. If: 
\begin{equation}
C_{i(N)}-C_{i(N-1)} < \eta
\end{equation}
where $\eta$ is a quite small threshold (generally it is smaller than 1e-4) that means it is almost impossible to generate an adversarial example through this $I_{wmi}$ due to the results we have observed from plenty of times of experiments. 

However, we simply set both the thresholds as 0 for the threshold noted in the previous paragraph and the threshold mentioned in Algorithm \ref{ISD} due to the difficulty to find the best values of these thresholds. The settings of these thresholds could be improvements in future work that will accelerate the adversarial example generation process and also enhance the quality of generated adversarial examples.

\section{Experiment}
\subsection{Basic Setting}
The CIFAR-10 dataset\footnote{https://www.cs.toronto.edu/~kriz/cifar.html} that contains 10 types (i.e., plane, car, bird, cat, deer, dog, frog, horse, ship, and truck) of image, and 1,000 images in each type, is used to test the proposed scheme. In addition, we simply utilize all images in the CIFAR-10 testing set as the watermark image dataset. That means  there are two CIFAR-10 testing sets exist in the experiment process according to our proposed framework, one is the image dataset and another is used as watermark image dataset. The configuration of the device used in the experiment displays in Table \ref{device}.

\begin{table}[!htbp]
    \centering
    \caption{Computer Configuration}
    \label{device}
    % \rowcolors{0}{}{gray!10}
    \begin{tabular}{*2l}
        \toprule
        Hardware / Software &  Configuration \\
        \hline
        \midrule
        CPU & Intel Xeon Silver 4210R 2.40 GHz\\
        GPU & NVIDIA GeForce RTX 3090 \\
        RAM & 320 GB \\
        Operating System & Windows 10 \\
        Python & 3.8.8 \\
        PyTorch & 1.9.1 \\
        \bottomrule
    \end{tabular}
\end{table}

We select eight typical DNNs to test our watermarking-based attacks, respectively VGG, GoogLeNet\cite{GoogLeNet}, ResNet, DenseNet121\cite{DenseNet},  MobileNet\cite{MobileNet}, and EfficientNet \cite{tan2019efficientnet}. We build and train these DNNs by Pytorch, referring to an excellent project on GitHub\footnote{https://github.com/kuangliu/pytorch-cifar}. Besides the performances of the trained models are shown in Table \ref{model}. 

\begin{table}[!htbp]
\centering
\caption{Model Performance on the CIFAR-10 dataset}
\label{model}
\begin{tabular}{|c|c|c|}
\hline
\diagbox{Model}{Performance}            & Accuracy & Loss  \\ \hline
VGG16       & 93.73\%       & 0.291 \\ \hline
VGG19       & 93.72\%       & 0.305 \\
\hline
GoogLeNet   & 95.28\%       & 0.151 \\ \hline
ResNet18    & 95.40\%       & 0.175 \\ \hline
ResNet50    & 95.52\%       & 0.194 \\ \hline
DenseNet121 & 95.48\%       & 0.173 \\ \hline
MobileNet   & 91.18\%       & 0.304 \\ \hline
EfficientNetB0   & 91.39\%  & 0.296 \\
\hline
\end{tabular}
\end{table}

It is noted that, we implement the experiments as the processes shown in Fig. \ref{framework}. First, we need to explain how to construct a watermark image dataset. In this work, we simply build the watermark dataset using the CIFAR-10 test set, which means the watermark dataset is totally the same as the image dataset.

\begin{figure}[!htbp]
  \centering
  \includegraphics[width=1\linewidth]{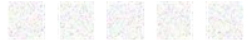}
  \caption{Noise Image Examples}
  \label{baseline}
\end{figure}

Besides, we design another baseline experiment to further verify the effectiveness of the proposed scheme. Concretely, we generate the baseline watermark dataset, which has the same number of images as the original watermark dataset, by adding random Gaussian noise to the pure white images as shown in Fig. \ref{baseline}.

\begin{figure*}[!htbp]
  \centering
  \includegraphics[width=\linewidth]{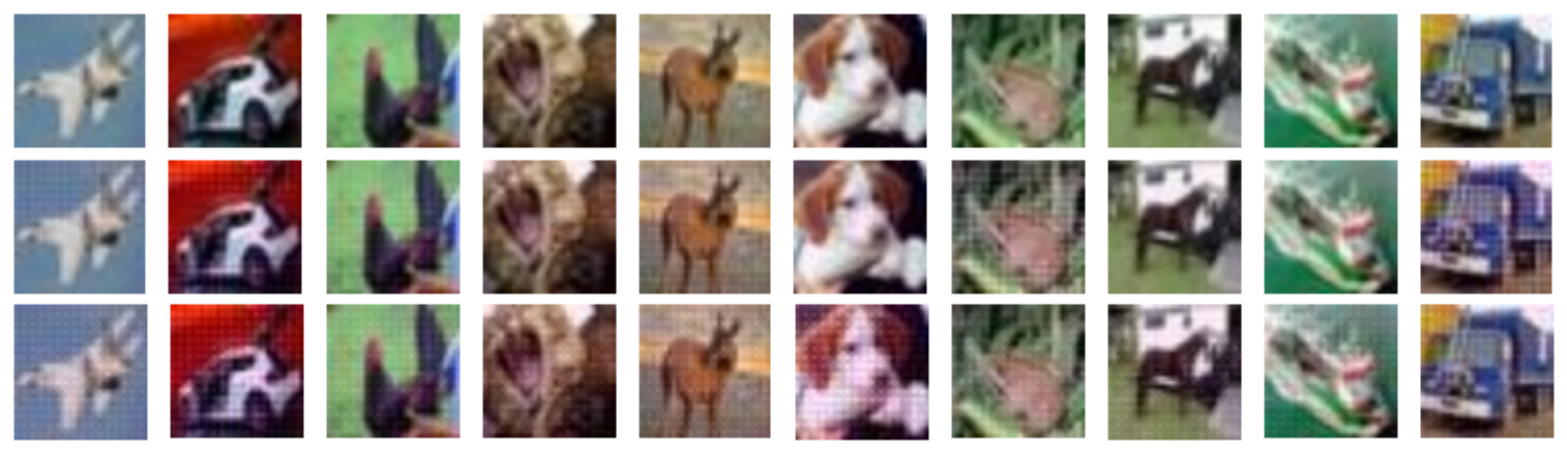}
  \caption{Comparison Between Original Images and Adversarial Examples}
  \label{show}
\end{figure*}

The generated adversarial examples are shown in Fig. \ref{show} where images in the first line are original images, and the images in the second and the third lines are the adversarial examples for DenseNet121 generated by the original watermark dataset and baseline watermark dataset respectively.

Next, we will illustrate the experimental results on the above two sets noted as general and baseline results from attack success rate and time consumption perspectives.

\subsection{General Experimental Result}

From Table \ref{e1-1}, we can see the entire attack success rate reaches 95.47\%, moreover, the highest success rate is 98.71\% on EfficientNetB0 and the lowest success rate is 89.77\% on VGG19. Also, the attack success rates are higher on EfficientNetB0, DenseNet121, MobileNet, ResNet18, and ResNet50, all exceeding or equal to 95.02\%. In addition, from the class perspective, most DNN models show better robustness in class-9 except ResNet18 and ResNet50. However, DenseNet121, MobileNet, VGG16, and VGG19 show better robustness in class-3.

\begin{table*}[!htbp]
\caption{Attack Success Rate on CIFAR-10 (Original)}
\centering
\begin{tabular}{|l|ccccccccccc|}
\hline
\diagbox{Model}{Rate}{Type}   & 0        & 1       & 2       & 3       & 4       & 5       & 6       & 7       & 8       & 9       & Total     \\[1ex] \hline
\makecell*[c]{VGG16}       & 98.60\%  & 95.70\% & 96.30\% & 92.20\% & 97.90\% & 99.00\% & 85.20\% & 89.80\% & 95.10\% & 82.70\% & \textbf{93.25\%} \\[1ex] 
\makecell*[c]{VGG19}      & 98.20\%  & 94.90\% & 96.40\% & 96.70\% & 93.80\% & 98.10\% & 85.60\% & 78.80\% & 87.70\% & 67.50\% & \textbf{89.77\%} \\[1ex] 
\makecell*[c]{GoogLeNet}   & 99.10\%  & 96.60\% & 98.80\% & 85.70\% & 99.10\% & 98.60\% & 92.50\% & 93.00\% & 91.50\% & 81.80\% & \textbf{93.67\%} \\[1ex] 
\makecell*[c]{ResNet18}    & 98.70\%  & 99.20\% & 99.20\% & 93.40\% & 99.60\% & 99.70\% & 96.40\% & 97.50\% & 98.60\% & 95.00\% & \textbf{97.73\%} \\[1ex] 
\makecell*[c]{ResNet50}    & 99.90\%  & 99.90\% & 98.80\% & 88.60\% & 99.60\% & 99.60\% & 97.20\% & 96.10\% & 99.90\% & 96.20\% & \textbf{97.58\%} \\[1ex] 
\makecell*[c]{DenseNet121} & 99.20\%  & 98.90\% & 98.50\% & 95.40\% & 99.70\% & 99.80\% & 93.80\% & 93.00\% & 90.70\% & 81.20\% & \textbf{95.02\%} \\[1ex] 
\makecell*[c]{MobileNet}  & 100.00\% & 99.60\% & 98.80\% & 98.70\% & 99.80\% & 99.80\% & 98.30\% & 96.90\% & 94.80\% & 94.00\% & \textbf{98.07\%} \\[1ex] 
\makecell*[c]{EfficientNetB0}  & 99.40\% & 99.40\% & 97.90\% & 98.90\% & 99.60\% & 99.40\% & 97.90\% & 97.50\% & 99.20\% & 97.90\% & \textbf{98.71\%} \\[1ex] 
\hline
\makecell*[c]{\emph{Avg}}  & 99.13\% &	98.02\%&	98.08\%&	93.70\%&	98.63\%&	 99.25\%&	93.36\%&	92.82\%& 94.68\%&	 87.03\%&	\emph{\textbf{95.47\%}} \\[1ex] 
\hline
\end{tabular}
\label{e1-1}
\end{table*}

As for time consumption, as shown in Fig. \ref{e1-2} the average speed to process each image is 1.17 seconds, but it is different in different DNN models. For example, the quickest speed is 0.58 seconds for processing each image on MobileNet and the slowest speed is 1.78 seconds on DenseNet121. Furthermore, for all classes (each class includes 1,000 images), the average total process speed of each class is under or equal to 2,603.04 seconds and the quickest speed achieves 515.26 seconds for class-1.

\begin{table*}[!htbp]
\caption{Time Consumption of Attack on CIFAR-10 (Original)}
\centering
\begin{tabular}{|l|rrrrrrrrrrrr|}
\hline
\diagbox{Model}{Seconds}{Type}   & 0        & 1       & 2       & 3       & 4       & 5       & 6       & 7       & 8       & 9       & Total  & Avg   \\[1ex] \hline
\makecell*[c]{VGG16}       & 426.52 & 1086.29 & 839.31  & 1678.56 & 652.86  & 528.19 & 1988.52 & 1960.66 & 1040.61 & 2613.14 & 12814.66 & \textbf{1.28} \\[1ex]  
\makecell*[c]{VGG19} & 524.41 & 1103.64 & 820.74  & 921.97  & 1374.02 & 579.05 & 1910.28 & 2641.25 & 1951.06 & 3913.03 & 15739.45 & \textbf{1.57} \\[1ex] 
\makecell*[c]{GoogLeNet}   & 692.21 & 1358.27 & 610.39  & 3461.34 & 756.10  & 716.98 & 1675.06 & 2187.40  & 2344.78 & 4049.78 & 17852.31 & \textbf{1.78} \\[1ex] 
\makecell*[c]{ResNet18}    & 661.44 & 414.74  & 373.00  & 1635.55 & 442.36  & 427.36 & 735.92  & 896.97  & 526.85  & 1200.17 & 7314.36  & \textbf{0.73} \\[1ex]  
\makecell*[c]{ResNet50}    & 348.36 & 447.04  & 662.52  & 2521.25 & 431.44  & 431.81 & 913.53  & 1250.69 & 454.90   & 1406.12 & 8867.66  & \textbf{0.88} \\[1ex] 
\makecell*[c]{DenseNet121} & 780.12 & 886.74  & 1158.17 & 2060.03 & 715.77  & 598.28 & 2188.97 & 3118.97 & 3641.35 & 5268.84 & 17592.21 & \textbf{1.75} \\[1ex]  
\makecell*[c]{MobileNet}  & 251.69 & 389.44  & 388.39  & 487.18  & 305.27  & 217.85 & 399.88  & 928.72  & 1151.61 & 1299.31 & 5819.34  & \textbf{0.58} \\[1ex]  
\makecell*[c]{EfficientNetB0}  & 437.37 & 716.16 & 841.71 & 929.93 & 557.63 & 683.82 & 1034.49 & 1387.15 & 531.97 & 1073.95 & 8194.18 & \textbf{0.81} \\[1ex] 
\hline 
\makecell*[c]{\emph{Avg}}  &  515.26& 	800.29& 	 711.77&   1711.97&   654.43&  522.91&  1355.83& 	 1796.47& 	 1455.39& 	2603.04 &  11774.27	& 
\emph{\textbf{1.17}} \\[1ex] \hline
\end{tabular}
\label{e1-2}
\end{table*}

According to the above results, we explore the relationship between the attack success rate and the time consumption. Generally, the DNN models with high attack success rates have shorter time consumption for processing each image except DenseNet121, i.e., EfficientNetB0 (98.71\% \& 0.81), MobileNet (98.07\% \& 0.58 seconds), ResNet18 (97.73\% \& 0.73 seconds), and ResNet50 (97.58\% \& 0.88 seconds).

\subsection{Baseline Experimental Result}

Based on the results displayed in Fig. \ref{e2-1}, for the baseline experiment, the average attack success rate is 88.82\% which decreases by 6.65\% compared with the results in general experiments. Furthermore, the highest attack success rate achieves 93.42\% on EfficientNetB0 and the lowest rate reaches 80.79\% on DenseNet121. Different from the general experimental results, we find several extreme phenomena. For example, in terms of class-3, the attack success rates are only 43.10\%, 45.70\%, and 50.40\% on ResNet50, DenseNet121, and GoogLeNet respectively. Similar to the above phenomena, this also happens to EfficientNetB0 for class-2, MobileNet and VGG19 for class-6, and ResNet18 and EfficientNetB0 for class-8. Additionally, the attack success rates still attain more than or equal to 91.31\% for ResNet18, ResNet50, MobileNet, and EfficientNetB0.

\begin{table*}[!htbp]
\caption{Attack Success Rate on CIFAR-10 (Baseline)}
\centering
\label{e2-1}
\begin{tabular}{|l|ccccccccccc|}
\hline
        \diagbox{Model}{Rate}{Type}     & 0       & 1       & 2       & 3       & 4       & 5       & 6       & 7       & 8       & 9       & Total            \\ \hline
\makecell*[c]{VGG16}       & 97.60\% & 94.00\% & 80.30\% & 77.60\% & 97.40\% & 96.30\% & 91.90\% & 96.20\% & 86.50\% & 79.50\% & \textbf{89.73\%} \\[1ex]
\makecell*[c]{VGG19}       & 96.90\% & 93.60\% & 71.80\% & 88.60\% & 80.60\% & 95.90\% & 54.90\% & 90.50\% & 84.20\% & 75.80\% & \textbf{83.28\%} \\[1ex] 
\makecell*[c]{GoogLeNet}   & 97.60\% & 95.00\% & 86.50\% & 50.40\% & 95.70\% & 91.10\% & 89.70\% & 95.80\% & 86.50\% & 90.90\% & \textbf{87.92\%} \\[1ex]
\makecell*[c]{ResNet18}    & 97.30\% & 98.10\% & 83.00\% & 88.50\% & 99.10\% & 98.40\% & 98.80\% & 98.20\% & 62.30\% & 89.40\% & \textbf{91.31\%} \\[1ex]
\makecell*[c]{ResNet50}    & 98.90\% & 99.60\% & 94.40\% & 43.10\% & 98.20\% & 97.10\% & 97.20\% & 98.10\% & 99.80\% & 98.00\% & \textbf{92.44\%} \\[1ex]
\makecell*[c]{DenseNet121} & 98.60\% & 99.00\% & 95.20\% & 45.70\% & 97.10\% & 93.70\% & 97.20\% & 97.20\% & 86.00\% & 91.90\% & \textbf{80.79\%} \\[1ex]
\makecell*[c]{MobileNet}  & 98.70\% & 98.60\% & 91.50\% & 90.60\% & 98.30\% & 99.20\% & 54.40\% & 99.00\% & 91.80\% & 94.60\% & \textbf{91.67\%} \\[1ex]
\makecell*[c]{EfficientNetB0}  & 97.70\% & 98.80\% & 73.90\% & 99.20\% & 98.90\% & 98.40\% & 99.60\% & 98.90\% & 68.00\% & 98.80\% & \textbf{93.42\%} \\[1ex] 
\hline
\makecell*[c]{\emph{Avg}} & 97.91\% & 97.08\% & 84.57\% & 72.96\% & 95.66\% & 96.26\% & 85.46\% & 96.73\% & 83.13\% & 89.86\% & \emph{\textbf{88.82\%}} \\[1ex]
\hline
\end{tabular}
\end{table*}

However, the change in average time consumption is not much compared to the general experiment, displayed in Table \ref{e2-2}, which only increases by 0.37 seconds for processing each image. For each class, the time consumption is normal except for class-3, especially, on DenseNet121, GoogLeNet, and ResNet50. As for EfficientNetB0, we observe that the time consumption for class-2 and class-8 is abnormal that is corresponding to the success rate results in Table \ref{e2-1}. In addition, the fastest processing speed is 453.99 seconds on class-0. 

\begin{table*}[!htbp]
\caption{Time Consumption of Attack on CIFAR-10 (Baseline)}
\centering
\label{e2-2}
\begin{tabular}{|l|rrrrrrrrrrrr|}
\hline
 \diagbox{Model}{Seconds}{Type}   & 0        & 1       & 2       & 3       & 4       & 5       & 6       & 7       & 8       & 9       & Total  & Avg   \\[1ex] \hline
\makecell*[c]{VGG16}      & 348.93   & 812.48   & 2232.53 & 2510.19  & 382.85   & 491.46   & 910.19   & 452.94  & 1681.98  & 2205.84  & 12029.39 & \textbf{1.20}         \\[1ex]
\makecell*[c]{VGG19}       & 470.23   & 751.35   & 3142.31 & 1341.35  & 2038.82  & 669.81   & 5130.22  & 1101.71 & 1759.79  & 2561.08  & 18966.67 & \textbf{1.89}  
\\[1ex] 
\makecell*[c]{GoogLeNet}   & 563.88   & 1003.58  & 2537.82 & 7774.98  & 898.92   & 1503.82  & 1856.22  & 851.88  & 2428.90   & 1639.24  & 21059.24 & \textbf{2.10}         \\[1ex]
\makecell*[c]{ResNet18}    & 441.32   & 294.87   & 1961.08 & 1534.82  & 197.13   & 276.58   & 247.98   & 308.02  & 4540.25  & 1347.38  & 11149.43 & \textbf{1.11}         \\[1ex]
\makecell*[c]{ResNet50}    & 293.81   & 203.11   & 838.10   & 7640.16  & 371.32   & 523.49   & 528.12   & 380.96  & 148.03   & 423.08   & 11350.18 & \textbf{1.13}         \\[1ex]
\makecell*[c]{DenseNet121} & 454.31   & 393.02   & 1132.77 & 10763.91 & 784.94   & 1587.54  & 732.62   & 729.54  & 3130.61  & 1854.13  & 21563.39 & \textbf{2.15}         \\[1ex]
\makecell*[c]{MobileNet}  & 218.50    & 234.88   & 894.76  & 1014.87  & 328.32   & 189.82   & 4690.10   & 189.24  & 902.11   & 602.96   & 9265.56  & \textbf{0.92}         \\[1ex]
\makecell*[c]{MobileNet}  & 251.69 & 389.44  & 388.39  & 487.18  & 305.27  & 217.85 & 399.88  & 928.72  & 1151.61 & 1299.31 & 5819.34  & \textbf{0.58} \\[1ex]  
\makecell*[c]{EfficientNetB0}  & 589.31 & 344.79 & 4293.21 & 305.89 & 310.79 & 400.94 & 196.42 & 289.78 & 5395.40 & 355.80 & 12482.33 & \textbf{1.24} \\[1ex] 
\hline
\makecell*[c]{\emph{Avg}}  &  453.99&  553.44&  2177.62&  4137.29&  664.79&  732.66&  1836.46&   654.09&  2642.33&  1263.12&  15460.69& \textit{\textbf{1.54}} \\ \hline
\end{tabular}
\end{table*}

The relationship between the attack success rate and the time consumption is similar to the results of the general experiment. ResNet18, ResNet50, MobileNet, and EfficientNetB0 still have high attack success rates while keeping less time consumption. 

In summary, the experimental results on both general and baseline groups illustrate the effectiveness and good performance of the proposed scheme. Furthermore, from the comparison between general and baseline experiments, we find it is important to select suitable watermark images which can enhance the performance of the proposed framework.

% \subsection{Baseline}
% We use the image that generated by the Gus noise as the baseline of our proposed scheme.
% \subsection{Defense}
% We use the framework proposed by Papernot et al.  \cite{papernot2018cleverhans} (NOT SURE or retraining the model with adversarial examples?) FUTURE work

\section{Discussion}

\subsection{Vision Transformer}
We also utilized Vision Transformer (ViT) \cite{dosovitskiy2020image} to test the proposed approach, which is implemented according to another GitHub project\footnote{https://github.com/kentaroy47/vision-transformers-cifar10}. The ViT model achieves 83.49\% accuracy rate and 1.148 loss on the CIFAR-10 testing set. 

Although the accuracy rate of the ViT model is not as high as the selected eight models, the attack success rates via the proposed framework are still not high. Specifically, for general experiments, the success rate is 66.66\% with 4.69 seconds of time consumption for each image. While for baseline experiments, the success rate only gets 53.27\% with 5.46 seconds of time consumption for each image.

Taking the general experimental results on the ViT model as an example, we observe that the attack success rates are high in a few classes (e.g., class-2 is 89.90\% and class-5 is 90.10\%), however, the attack success rates for most classes are between 52.90\% to 75.10\%. Also, we notice that the lowest attack success rate is 35.90\% in class-9.

Based on the above experiments, we consider that the main reason why the proposed framework has worse experimental results on ViT model is related to the structural design which is different from the eight models we selected. Thus, further research can explore the relationships between the proposed scheme and specific model structures. Besides, we also come up with the enhancement for the current scheme in the following sections.

\subsection{Improved DCT-based Watermarking}

Similar to Algorithm \ref{DWT}, we propose an enhanced DCT-based watermarking algorithm as illustrated in Algorithm \ref{DCT}. However, after a number of attempts, we did not find a suitable set of parameters for it to generate adversarial examples. 

\begin{algorithm}[htbp!]
\DontPrintSemicolon
\KwIn{$I_i$, $I_{wmi}$}
\KwOut{$I_{i(N)}$}

\For {$i = 1, 2, ...$} {
    \For{$j = 1, 2, ..., N$} {
        read and resize $I_{i(j)}$ and $I_{wmi}$;\\

        separate $R$, $G$ and $B$ channels of $I_{i(j)}$ and $I_{wmi}$;\\
        
        \For{$c = \{R, G, B\}$} {
            $I_{i(j)}^{c}$
            $T^c  \leftarrow  DCT(I_{i(j)}^{c})$;\\
            $T_{wm}^c  \leftarrow  DCT(I_{wmi}^{c})$;\\
            $E^c  \leftarrow  T^c + S_c \times T_{wm}^c$;\\
            
            $E^c  \leftarrow  IDCT(E^c)$;\\
        }
        
        $I_{i(j)}  \leftarrow  cat(3,E^{R},E^{G},E^{B})$;\\
    }
}
\Return $I_{i(j)}$
\caption{Improved DCT-based Watermarking}
\label{DCT}
\end{algorithm}

\begin{figure}[!htbp]
  \centering
  \includegraphics[width=0.65\linewidth]{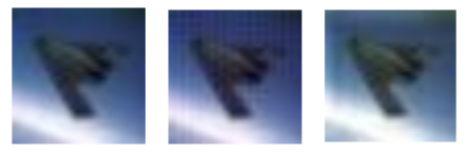}
  \caption{Comparison of the Original Image and Different Adversarial Examples (from left to right are the original image, the DWT adversarial example, and the DCT adversarial example)}
  \label{o-dwt-dct}
\end{figure}

Specifically, we find that in order to obtain better results on adversarial example generation for Algorithm \ref{DCT}, the embedding strength parameters need to be smaller than those in Algorithm \ref{DWT}, and the number of embedding times has to be more compared with Algorithm \ref{DWT}. The experimental results are unsatisfying because it takes quite more time to generate adversarial examples compared with Algorithm \ref{DWT}, though the perturbations seem to be less on generated images as shown in Fig. \ref{o-dwt-dct}. One of the parameter sets we test is shown below:
\begin{equation}\label{RGB_P_DWT}
    \begin{cases}
    S_r:S_g:S_b = 0.010:0.020:0.005,\\
    N_{max} = 10.
    \end{cases}
\end{equation}.

We believe it is possible to find an appropriate set of parameters to efficiently generate adversarial examples through the above DCT-based watermarking algorithm. Due to the difficulty of finding good parameters, we consider designing a mechanism to automatically seek out parameters for our proposed scheme in future work according to the feedback from the results, which we will further discuss in the following section.

\subsection{Parameter Selection Mechanism}
For both DWT-based and DCT -based watermarking algorithms, in order to find appropriate parameters (i.e., embedding strength and times), an efficient method is to design a mechanism to search them automatically according to the specific rules. Although we have yet designed and implemented this mechanism, we consider there is two feedback that should be included that are: 
\begin{enumerate}
    \item Color information of the original image and the generated image.
    \item Classification result and confidence of the original image and the generated image.
\end{enumerate}

We will introduce the basic idea about how to use the above components to build the mentioned mechanism. For preventing the changes in the images from being aware by HVS, we are able to choose the specific proportion of RGB for each image by the color information (e.g., color distribution). 

Then, we can set specific thresholds in Algorithm \ref{ISD} that we have yet set in our current experiments according to the changes in classifications and confidences.

\subsection{Experiment Discussion and Enhancement}
We can find some interesting results from the experiments, we select two examples to discuss. First, the attack success rates are high on ResNet18, ResNet50, MobileNet, and EfficientNetB0 for both general and baseline experiments, which might be caused by the natural structure of the specific models. That means the proposed scheme might work well on some DNN models but not all of them. To further understand the reason why this happens, we need to explore the relationship between the influence on original images due to embedding the watermark image and DNN model structures.

Second, the attack success rates are low in specific class on specific model for both general and baseline experiments (e.g., class-9 on VGG19). We consider there are three possible reasons:
\begin{enumerate}
    \item The bias of the trained DNN model in different classes.
    \item The original image has the ability to resist this kind of attack to an extent.
    \item The selected watermark image is unsuitable because the process is random.
\end{enumerate}
However, to answer which reason really affects the results or has the most effects on the results, we need to further study.

In addition, we have four initial ideas for improving the current scheme. First, it is possible to embed different watermark images in each round according to the attention maps of the original image and the watermark image, which may lead to the biggest influence in each round on the original image for generating the adversarial example.

Second, we can craftily design a specific watermark image or a set of special watermark images that can be easily used to generate adversarial examples after they are embedded into the original images. The difficulty of this method is finding a way to generate these watermark images. For example, we can try to generate many watermark images as we build the baseline watermark image dataset, and test each one, but it will cost lots of time.

Third, we can devise an approach similar to the method mentioned in \cite{9534119}, which preprocess the watermark image before it is embedded. After the special process, the watermark image may have more information and easier to be utilized for adversarial example generation.

Fourth, based on the attention maps, we can divide the image into several blocks and embed the focus part of the watermark image shown in the corresponding attention map into the focus part of the original image (produce maximum overlap), then recombine the original image.

% DCT does not work well, maybe we need to find a specific noise for it? or find the suitable parameters?
% Also, we can the potential of the proposed method that can be applied in different domains not only image but also audio \cite{10.1145/3472634.3474080}
% 1. special watermark
% 2. the nature of the model structure
% The results in experiments might cause by the watermark imgae that embedded.?
% DCT does not play well but we show one result of it
% For similar DCT method \cite{9534119} may help which proposed after our paper first version on
% arXiv 
% they use specific image, but we use any possible images
% Other algorithm to embedding watermarks
% divide one image to many blocks then random embedding watermark image into some block?
% extract middle layer image as the watermark image.
% it is possible to change the watermark dataset to a specific image that can .... generate adversarial examples easily

\section{Conclusion}
In this paper, we propose a framework for generating adversarial examples based on embedding watermark images with a special set of parameters. Specifically, we propose an improved DWT-based watermarking algorithm as the core approach in the proposed framework. Besides, we design a baseline experiment implemented by the images generated by random noises to prove the effectiveness of the proposed scheme. The experimental results illustrate that the proposed scheme is feasible and efficient in that the attack success rate achieves 95.47\% and 1.17 seconds for processing each image. In addition, we also propose a similar DCT-based watermarking algorithm without suitable parameters, which can be further explored in future work.

%--------------------------------------------------------

% \begin{acks}
% The research was financially supported by National Natural Science Foundation of China (No. 61972366), the Provincial Key Research and Development Program of Hubei (No. 2020BAB105), and the Henan Key Laboratory of Network Cryptography Technology (No. LNCT2020-A01). K.-K.R. Choo was funded only by the Cloud Technology Endowed Professorship.
% \end{acks}

\bibliographystyle{IEEEtran}
\bibliography{sample-base}

\end{document}